\documentclass[10pt,journal,compsoc]{IEEEtran}

\ifCLASSOPTIONcompsoc
  \usepackage[nocompress]{cite}
\else
  \usepackage{cite}
\fi
\ifCLASSINFOpdf
\else
\fi

\usepackage{epsfig}
\usepackage{graphicx}
\usepackage{grffile} %
\usepackage{amsmath}
\usepackage{amssymb}
\usepackage{booktabs, makecell, tabularx}
\usepackage{subfig}
\usepackage[numbers,sort]{natbib}
\usepackage{colortbl}
\usepackage{algpseudocode}
\usepackage{algorithm}

\usepackage{setspace}

\usepackage{rotating}
\usepackage[export]{adjustbox}
\usepackage{xcolor}
\usepackage{url}
\usepackage{ragged2e}
\usepackage{xspace}
\usepackage{setspace}
\usepackage{twoopt}
\usepackage{anyfontsize}
\usepackage{color,soul}

\usepackage{tikz}

\usetikzlibrary{calc,fit}

\newcommandtwoopt\Textbox[6][2.5cm][2cm]{%
\begin{tikzpicture}[remember picture, overlay]
  \coordinate (aux) at ([xshift=#1]#4);
  \node[inner ysep=4.7pt,yshift=0.4ex,draw=#6,thick,
    fit=(#3) (aux),baseline] 
    (box) {};
  \node[text width=#2,anchor=north east,
    font=\footnotesize,align=right] 
    at (box.north east) {#5};
\end{tikzpicture}%
}
\DeclareCaptionSubType*{algorithm}

\DeclareCaptionLabelFormat{alglabel}{Alg.~#2}

\newcommand{\figref}[1]{Fig.~\ref{#1}}
\newcommand{\figureref}[1]{Figure \ref{#1}}
\newcommand{\secref}[1]{Section~\ref{#1}}
\newcommand{\algoref}[1]{Algorithm~\ref{#1}}
\newcommand{\eqnref}[1]{Eq.~\eqref{#1}}
\newcommand{\tabref}[1]{Table~\ref{#1}}

\newcommand{\pardev}[2]{\frac{\partial {#1}}{\partial {#2}}}

\def\eg{e.g.\xspace} 
\def\ie{i.e.\xspace} 
\def\cf{cf.\xspace} 

\def\vs{vs.\xspace}
\def\wrt{wrt.\xspace}

\newcommand{\boldparagraph}[1]{\vspace{0.2cm}\noindent{\bf #1:}}

\newcommand{\netparam}{\sigma}
\newcommand{\point}{\mathbf{x}}

\newcommand{\function}[1]{{#1}_{\netparam_{#1}}}
\newcommand{\mfunction}[1]{\mathbf{#1}_{\netparam_{#1}}}

\newcommand{\occ}{o}
\newcommand{\loss}{\mathcal{L}}

\newcommand{\bone}{\boldsymbol{B}}

\newcommand{\jac}{\mathbf{J}}

\newcommand{\name}{Fast-SNARF\xspace}
\newcommand{\snarf}{SNARF\xspace}

\begin{document}
\title{Fast-SNARF: A Fast Deformer for Articulated Neural Fields}

\author{\fontsize{10}{13.6}\selectfont{Xu~Chen*\textsuperscript{{1,2}} 
        Tianjian~Jiang*\textsuperscript{{1}}
        Jie~Song\textsuperscript{{1}}
        Max~Rietmann\textsuperscript{{3}}
        Andreas~Geiger\textsuperscript{{2,4}}
        Michael~J.~Black\textsuperscript{{2}}
        Otmar~Hilliges\textsuperscript{{1}}\thanks{*Xu Chen and Tianjian Jiang contributed equally.} \\
        \vspace{0.2cm}\textsuperscript{{1}}ETH Z{\"u}rich \quad \textsuperscript{{2}}Max Planck Institute for Intelligent Systems, T{\"u}bingen \quad \textsuperscript{{3}}NVIDIA \quad \textsuperscript{{4}}University of T{\"u}bingen
        }
        }%

\markboth{Journal of \LaTeX\ Class Files,~Vol.~14, No.~8, August~2015}%
{Shell \MakeLowercase{\textit{et al.}}: Bare Advanced Demo of IEEEtran.cls for IEEE Computer Society Journals}
\IEEEtitleabstractindextext{%
\vspace{-1em}
\begin{abstract}
\justifying
Neural fields have revolutionized the area of 3D reconstruction and novel view synthesis of \emph{rigid} scenes. 
A key challenge in making such methods applicable to \emph{articulated} objects, such as the human body, is to model the deformation of 3D locations between the rest pose (a canonical space) and the deformed space.
We propose a new articulation module for neural fields, \name, which finds accurate correspondences between canonical space and posed space via iterative root finding. 
\name is a drop-in replacement in functionality to our previous work, \snarf, while significantly improving its computational efficiency. 
We contribute several algorithmic and implementation improvements over SNARF, yielding a speed-up of $150\times$. 
These improvements include voxel-based correspondence search, pre-computing the linear blend skinning function, and an efficient software implementation with CUDA kernels.
\name \ enables efficient and simultaneous optimization of shape and skinning weights given deformed observations without correspondences (\eg \ 3D meshes).
Because learning of deformation maps is a crucial component in many 3D human avatar methods and since \name provides a computationally efficient solution, we believe that this work represents a significant step towards the practical creation of 3D virtual humans.

\end{abstract}
\vspace{-0.5em}
}

\maketitle

\IEEEpeerreviewmaketitle

\ifCLASSOPTIONcompsoc
\IEEEraisesectionheading{\section{Introduction}\label{sec:introduction}}
\else
\section{Introduction}
\fi
3D avatars are an important building block for many emerging applications in the metaverse, AR/VR and beyond. 
To this end, an algorithm to reconstruct and animate non-rigid articulated objects, such as humans, accurately and quickly is required. This challenging task requires modeling the 3D shape and deformation of the human body -- a complex, articulated, non-rigid object. For such techniques to be widely applicable, it is paramount that algorithms do not require manually provided annotations nor that subjects appear in a-priori known poses. Therefore, inferring the transformation that 3D locations undergo between the posed observation space and some canonical space is the key challenge to attain a model that can be animated.

\begin{figure}[t]
    \centering
    \includegraphics[width=\linewidth,trim=0 0 0 0, clip]{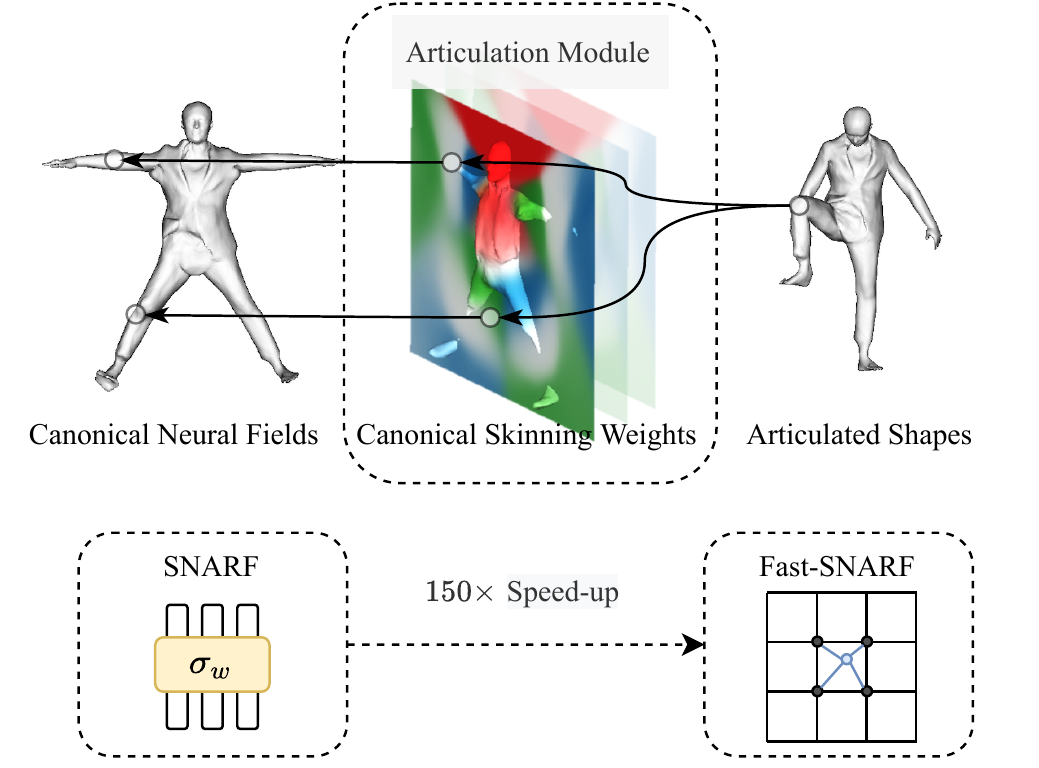}
    \captionof{figure}{\textbf{\name \ for Articulated Neural Fields.} \name \ finds accurate correspondences between canonical space and posed space while being 150$\times$ faster than our previous method SNARF~\cite{Chen2021ICCV}. \name \ enables optimizing shape and skinning weights given deformed observations without correspondences (\eg \ 3D meshes).
    \vspace{-2em}
    }
    \label{fig:teaser}
\end{figure}

Static shape modeling has recently seen much progress with the advent of neural fields~\cite{Mescheder2019CVPR,Park2019CVPRDeepsdf,Mildenhall2020ECCV, mueller2022instant}.
Such representations are promising due to their ability to represent complex geometries of arbitrary topology at arbitrary resolution, by leveraging multi layer perceptrons (MLPs) to encode spatial quantities of interest (\eg \ occupancy probabilities) in 3D space. 
Recent work~\cite{mueller2022instant} has further achieved fast reconstruction and real-time view synthesis of rigid scenes with high quality. However, to enable fast \emph{non-rigid} reconstruction and realistic \emph{animation} of articulated objects, a robust and fast articulation module is needed.

Articulation of neural fields is typically modeled via deformation of 3D space, which warps neural fields from a rest pose (canonical space) into any target pose (posed space), leveraging dense deformation fields. Several techniques have been proposed to construct such deformation fields. Building upon traditional mesh-based linear blend skinning (LBS)~\cite{James:1995}, several works~\cite{Jeruzalski2020ARXIV,Mihajlovic2021CVPR,Saito2021CVPR,Peng2021ICCV,Tiwari2021ICCVNeuralGIF} learn dense skinning weight fields in \emph{posed space} and then derive the deformation fields via LBS. While inheriting the smooth deformation properties of LBS, the resulting skinning weight fields cannot generalize to unseen poses, because the are \emph{pose-dependent} and changes in pose lead to drastic changes to the spatial layout of the deformation field. These changes not been observed at training time for unseen poses. Another line of work approximates the mapping as piece-wise rigid transformations~\cite{Deng2020ECCV,2021narf}, which suffers from discontinuous artifacts at joints. The mapping could also be approximated based on a skinned base mesh~\cite{Huang2020CVPR}, which can lead to inaccuracies due to the mismatch between the base mesh and the actual shape and suffers from erroneous nearest neighbor associations in regions with self-contact.

Our recent work, SNARF~\cite{Chen2021ICCV}, overcomes these problems by design in that it learns a skinning weight field in canonical space which is \emph{pose-independent}. 
This formulation allows for natural deformation due to the smooth deformation properties of LBS and generalizes to unseen poses because of  the pose-independent canonical skinning weights. Furthermore, in contrast to previous methods~\cite{Jeruzalski2020ARXIV,Mihajlovic2021CVPR,Saito2021CVPR,Peng2021ICCV,Tiwari2021ICCVNeuralGIF}, pose-independent skinning weights can be learned unsupervised, \ie without the need for ground-truth skinning weights or other forms of annotations.

However, a major limitation of SNARF is the algorithm's computational inefficiency. While learning a canonical skinning weight field enables generalization, the deformation from posed to canonical space is defined implicitly, and hence can only be determined numerically via iterative root finding. The efficiency of the operations at each root finding iteration play a critical role in the speed of the overall articulation module. Therefore, computationally expensive operations in SNARF, such as computing LBS and evaluating the skinning weight field, parameterized by an MLP, lead to prohibitively slow speed -- learning an animatable avatar from 3D meshes takes 8 hours on high-end GPUs.

In this paper, we propose \name, an articulation module that is fast yet preserves the accuracy and robustness of \snarf. We achieve this by significantly reducing the computation at each root finding iteration in the articulation module. First, we use a compact voxel grid to represent the skinning weight field instead of an MLP. The voxel-based representation can replace MLPs without loss of fidelity because the skinning weight field is naturally smooth, and is pose-independent in our formulation. In addition, exploiting the linearity of LBS, we factor out LBS computations into a pre-computation stage without loss of accuracy. As a result, the costly MLP evaluations and LBS calculations in SNARF are replaced by a single tri-linear interpolation step, which is lightweight and fast. Together with a custom CUDA kernel implementation, \name can deform points with a speed-up of 150x \wrt \snarf (from 800ms to 5ms) without loss of accuracy.

In our experiments we follow the setting of \snarf and learn an animatable avatar, including its shape and skinning weights, from 3D scans in various poses, represented by a pose-conditioned occupancy field parameterized by an MLP. The overall inference and training speed, including both articulation and evaluation of the canonical shape MLP, is increased by $30\times$ and $15\times$ respectively. Note that the speed bottleneck is shifted from articulation (in SNARF) to evaluating the canonical shape MLP (in \name). \name is also faster than other articulation modules and is significantly more accurate, as we show empirically.
While we focus on learning occupancy networks, \name may be interfaced with other neural fields in the same manner that \snarf and variants have been utilized \cite{li2022tava,ARAH:2022:ECCV, zheng2022avatar,jiang2022selfrecon}.

We hope \name \ will accelerate research on articulated 3D shape representations
and release the code on our project webpage \footnote{\url{https://github.com/xuchen-ethz/fast-snarf}} to facilitate future research.

\boldparagraph{Relation to SNARF~\cite{Chen2021ICCV}} This paper is an extension of SNARF~\cite{Chen2021ICCV}, a conference paper published at ICCV '21 which models articulation of neural fields. This paper addresses the main limitation of SNARF, \ie \ its computational inefficiency via a series of algorithmic and implementation improvements described in \secref{sec:fast_snarf}. We provide a speed and accuracy comparison of \name \ with SNARF and other baseline methods, and thorough ablations in \secref{sec:exp}. %

\section{Related Work}
\subsection{Rigid Neural Fields}

Neural fields have emerged as a powerful tool to model complex rigid shapes with arbitrary topology in high fidelity by leveraging the expressiveness of neural networks. These neural networks regress the distance to the surface~\cite{Park2019CVPRDeepsdf}, occupancy probability~\cite{Mescheder2019CVPR}, color~\cite{Oechsle2019ICCV} or radiance~\cite{Mildenhall2020ECCV} of 3D points. Conditioning on local information such as 2D image features or 3D point cloud features produces more detailed reconstructions~\cite{Chibane2020CVPR, He2020NIPS, Peng2020ECCV, Saito2019ICCV, Saito2020CVPR} than using global features. Such representations can be trained with direct 3D supervision, e.g.~ground truth occupancy or distance to the surface, or can be trained indirectly with raw 3D points clouds~\cite{Atzmon2020CVPR, Gropp2020ICML, Saito2021CVPR} or 2D images~\cite{Mildenhall2020ECCV, Niemeyer2020CVPR, Sitzmann2019NIPS, Yariv2020NIPS}.

\boldparagraph{Fast Rigid Neural Fields}
One major limitation of neural field representations are their slow training and inference speeds, mainly due to the fact that multiple evaluations of deep neural networks are necessary to generate images and each of these evaluations is time-consuming. 
Several works have recently been proposed to improve the training ~\cite{Liu20neurips_sparse_nerf, takikawa2021nglod, yu_and_fridovichkeil2021plenoxels, SunSC22, mueller2022instant, Chen2022ECCV} and inference speed~\cite{reiser2021kilonerf, garbin2021fastnerf, hedman2021snerg, yu2021plenoctrees}. The core idea is to leverage explicit representations~\cite{Peng2020ECCV}, such as voxel grids or hash tables, to store features for a sparse set of points in space. The dense field can then be obtained by interpolating sparse features and by decoding the features using neural networks. 
Instead of point locations, these networks take features as input, which are more informative, enabling the network to be shallow and hence more computationally efficient.
However, the underlying explicit representations have a fixed spatial layout which limits these methods to rigid shapes. 

Our proposed articulation module can deform rigid neural fields to enable non-rigid animation at inference time and enable learning from deformed observations during training. Importantly, our module runs at a comparable speed to recent fast rigid neural field representations (\eg \cite{mueller2022instant}) and is thus complementary to advancements made in accelerating neural fields.

\subsection{Articulation of Neural Fields}
Recently, several articulation algorithms for neural fields have been proposed. 
These methods serve as a foundation for many tasks such as generative modeling of articulated objects or humans~\cite{Corona2021CVPRSMPLicit, chen2022gdna, hong2022avatarclip, noguchi2022unsupervised, bergman2022gnarf, zhang2022avatargen}, and reconstructing animatable avatars from scans~\cite{Deng2020ECCV,Saito2021CVPR,Mihajlovic2021CVPR,Chen2021ICCV,Tiwari2021ICCVNeuralGIF, lin2022fite, Mihajlovic:CVPR:2022}, depth~\cite{Palafox2021ICCV, wang2021metaavatar, dong2022pina}, videos~\cite{2021narf, peng2021animatable, liu2021neural, chen2021animatable, zheng2022avatar,  peng2022animatable, jiang2022selfrecon, weng_humannerf_2022_cvpr, ARAH:2022:ECCV, li2022tava, jiang2022neuman} or a single image~\cite{Huang2020CVPR,he2021arch++,xiu2022icon}. 

\boldparagraph{Part Based Models} One option is to model articulated shapes as a composition of multiple parts~\cite{Deng2020ECCV, 2021narf, Mihajlovic:CVPR:2022}. Rigidly transforming these parts according to the input bone transformations produces deformed shapes. While preserving the global structure after articulation, the continuity of surface deformations is violated, causing artifacts at the intersections of parts. Moreover, inferring the correct part assignment from raw data is challenging and typically requires ground-truth supervision.

\boldparagraph{Backward Skinning} 
Another line of work~\cite{Jeruzalski2020ARXIV,Saito2021CVPR,Mihajlovic2021CVPR, Tiwari2021ICCVNeuralGIF} proposes to learn skinning weight fields in deformed space and then derive the backward warping field using LBS to map points in deformed space to canonical ones. Such methods are straightforward to implement but inherently suffer from poor generalization to unseen poses. Backward deformation fields are defined in deformed space and, hence, inherently deform with the pose. Thus, the network must memorize deformation fields for different spatial configurations, making it difficult to generate deformations that have not been seen during training. Learning such pose-dependent skinning weight fields is also challenging, thus existing methods often rely on strong supervision via ground-truth skinning weights. Moreover, due to the varying spatial configuration, such pose-dependent skinning weights cannot be modeled using acceleration data structures such as explicit voxel grids.

\boldparagraph{Forward Skinning} Learning the skinning weights in canonical space instead of deformed space is a natural way to resolve the generalization issue. However, deriving the mapping from deformed to canonical points with canonical skinning weights is not straightforward, because the skinning weights of the deformed query points are unknown. Thus, SNARF~\cite{Chen2021ICCV} attains this mapping using an iterative root finding formulation, which finds the canonical points that are forward-skinned to the deformed query location. This formulation enables the articulation of neural fields into arbitrary poses, even those unseen during training. The pose-independent canonical skinning weights can be learned unsupervised without the need for ground-truth skinning weights. Moreover, multiple canonical correspondences can be found using such methods, which is important to handle self-contact. This forward skinning formulation has already found widespread use in many tasks, such as generative modeling~\cite{chen2022gdna}, or personalized avatar reconstruction from scans~\cite{lin2022fite}, depth \cite{dong2022pina}, or images~\cite{jiang2022selfrecon, zheng2022avatar, ARAH:2022:ECCV, li2022tava}.

However, one major limitation of this formulation is its slow speed due to the expensive computation at each root finding iteration. The original SNARF model relies on an MLP to parameterize the skinning weight field. At each root finding iteration, SNARF requires evaluating the MLP to compute LBS weights, which is time-consuming. 
This limitation is further amplified when combining forward skinning with rendering algorithms that require many queries along many rays (\cf ~\cite{corona2022lisa}). To reduce computation time, existing methods~\cite{jiang2022selfrecon, ARAH:2022:ECCV} use an explicit mesh to tighten the search space of root finding. However, these methods introduce the overhead of mesh extraction and still require days of training time to learn avatars from images.

We address this problem by using a voxel-based parameterization of the skinning weight field and by factoring out the LBS computation into a pre-computation stage. Since \name does not require mesh extraction in the training loop and is therefore more versatile and much faster to train than those that rely on meshes (\eg~\cite{jiang2022selfrecon}) (minutes \vs \ days). Our method also enables learning the skinning weights. 

\section{Differentiable Forward Skinning}
\label{sec:snarf}

\begin{figure*}
    \centering
    \includegraphics[width=\textwidth]{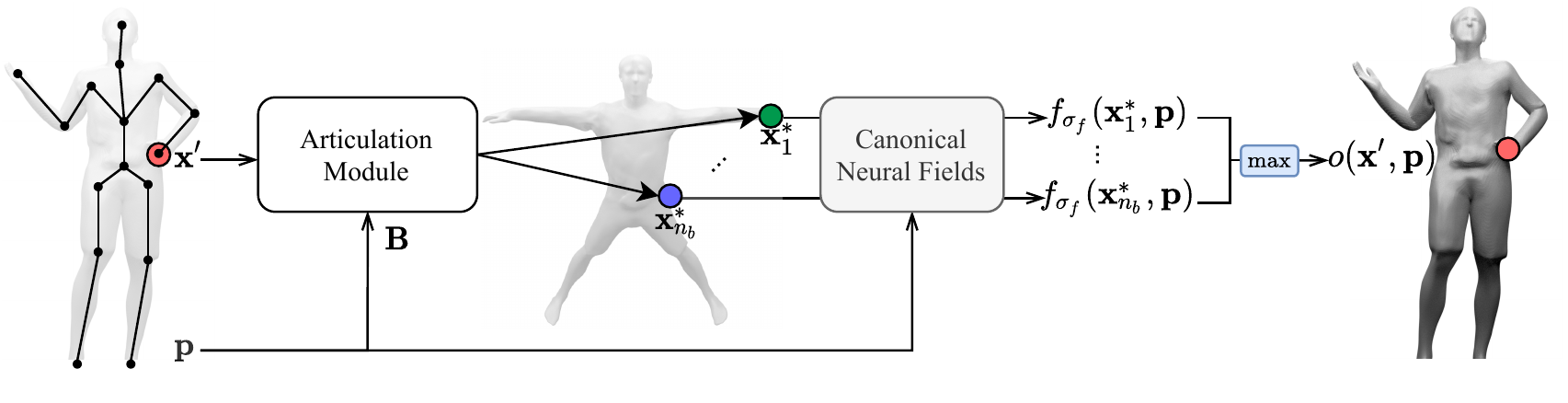}
    \vspace{-2.5em}
    \caption{\textbf{General Framework for Articulated Neural Field Representations.} 
    Given a query point in deformed space $\point'$ and the input pose (represented as joint angles $\mathbf{p}$ and 6D transformations $\mathbf{B}$), an articulation module first finds its canonical correspondences $\point^*$. The canonical shape representation $\function{f}$ then outputs the occupancy probabilities or densities at $\{\point^*\}$ which are finally aggregated to yield the occupancy probability or density of the query point $\point'$. 
    \vspace{-1em}
    }
    \label{fig:pipeline}
\end{figure*}

In this section, we briefly summarize the differentiable forward skinning approach proposed in SNARF~\cite{Chen2021ICCV}. We then discuss Fast-SNARF in \secref{sec:fast_snarf}.

\boldparagraph{General Pipeline} \figureref{fig:pipeline} illustrates the general pipeline for modeling articulated neural fields. 
Given a query point in posed space, an articulation module first finds its correspondences in canonical space according to the input body pose.
Then the canonical shape %
properties are evaluated at the correspondence locations. When multiple correspondences exist, multiple values of these properties are predicted and aggregated into one value as the final output.

\boldparagraph{Canonical Neural Fields} %
Canonical shape %
properties can be modeled using any coordinate-based representation, \eg \ occupancy fields~\cite{Mescheder2019CVPR} or radiance fields~\cite{Mildenhall2020ECCV}.
For convenience, we follow SNARF and use occupancy fields as an example. 
The occupancy field in SNARF~\cite{Chen2021ICCV} is defined as
\begin{align} 
\function{f}: \mathbb{R}^3 \times \mathbb{R}^{n_p}&\rightarrow [0,1], \\
\point, \mathbf{p} &\mapsto \occ .
\end{align}
Here $\function{f}$ is the occupancy field that predicts the occupancy probability $\occ$ for any canonical point $\point$. The parameters of the occupancy field are denoted as $\sigma_f$. It can be optionally conditioned on the articulated pose $\mathbf{p}$ to model pose-dependent local deformations such as clothing wrinkles.

\boldparagraph{Neural Blend Skinning}
In SNARF, the articulation is modeled using LBS. To apply LBS to continuous neural fields, a skinning weight field in canonical space is defined as:
\begin{align} 
\mfunction{w}: \mathbb{R}^3 \rightarrow \mathbb{R}^{n_b} ,
\end{align}
where $\sigma_w$ are the parameters and $n_b$ denotes the number of bones. In SNARF, this field is parameterized as an MLP. However, any other coordinate-based representation can be used instead. Given the skinning weights $\mathbf{w}$ of a 3D point $\point$ and the bone transformations $\bone=\{\bone_1,\dots,\bone_{n_b}\}$ ($\bone_{i}\in SE(3)$) that correspond to a particular body pose $\mathbf{p}$, the 6D transformation $\mathbf{T}(\point)\in \mathbb{R}^{3\times4}$ of a canonical point is determined by the following convex combination:
\begin{align}
\mathbf{T}(\point) = \sum_{i=1}^{n_{\text{b}}}{w}_{\sigma_{w},i}(\point) \cdot \bone_i
. 
\label{equ:lbs}
\end{align}

The deformed point corresponding to the canonical point is then computed as
\begin{align}
\point' = \mathbf{d}_{\sigma_w}(\point,\bone) = \mathbf{T}(\point) \cdot \point. 
\label{equ:lbs2}
\end{align}

\begin{figure}[t!]
    \centering
    \includegraphics[width=\linewidth,trim=0 0 0 0,clip]{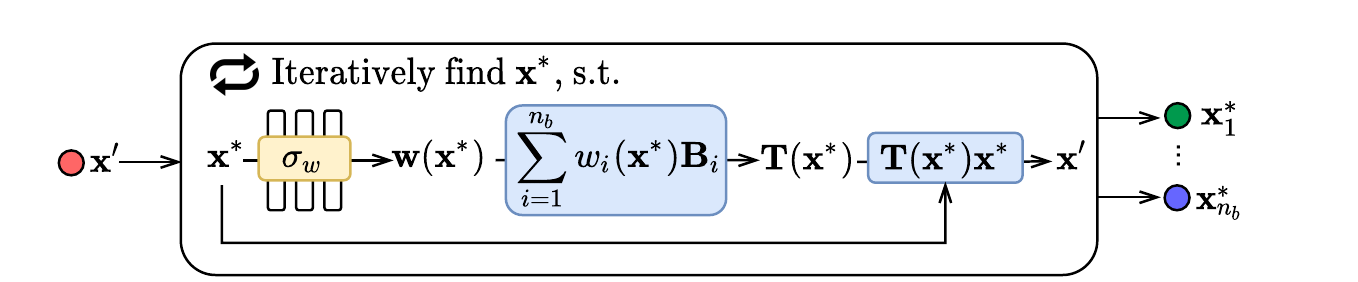}
    \caption{\textbf{MLP-based Forward Skinning~(SNARF).} 
    Given a point in deformed space $\point'$, SNARF finds its canonical correspondences $\point^*$ that satisfy the forward skinning equation \eqref{equ:lbs2} via root finding. Multiple correspondences can be reliably found by initializing the root finding algorithm with multiple starting points derived from the bone transformations. 
    }
    \label{fig:snarf}
\end{figure}

\boldparagraph{Correspondence Search} 
The canonical skinning weight field and \eqnref{equ:lbs2} define the mapping from canonical points to deformed ones, \ie \
$\point \rightarrow \point'$. However, generating posed shapes requires the inverse mapping, \ie \
$\point' \rightarrow \point$, which is defined implicitly as the root of the following equation:
\begin{align}
\mathbf{d}_{\sigma_w}(\point,\bone) - \point' = \mathbf{0} .
\label{equ:root_equ}
\end{align}
The roots of this equation cannot be analytically solved in closed form. 
Instead, the solution can be attained numerically via standard Newton or quasi-Netwon optimization methods, which iteratively find a location $\point$ that satisfies \eqnref{equ:root_equ} (see \figref{fig:snarf}):
\begin{align}
\point^{k+1} &= \point^{k} -(\jac^k)^{-1}\cdot ( \mathbf{d}_{\sigma_w}(\point^k, \bone) - \point' ) .
\label{equ:broyden}
\end{align}
Here $\jac$ is the Jacobian matrix of $\mathbf{d}_{\sigma_w}(\point^k, \bone) - \point'$. To avoid computing the Jacobian at each iteration, Broyden's method \cite{Broyden1965BOOK} and low-rank approximation $\Tilde{\jac}$ of $\jac^{-1}$ is used.

\boldparagraph{Handling Multiple Correspondences}
Multiple roots, denoted by the set $\{\point_i^*\}$, might exist due to self-contact where multiple canonical correspondences of one deformed point exist (see green and blue points in \figref{fig:pipeline}). Multiple roots are found by initializing the optimization-based root finding procedure with different starting locations and exploiting the local convergence of the optimizer. The initial states $\{\point_i^0\}$ are thereby obtained by transforming the deformed point $\point'$ rigidly to the canonical space for each of the $n_{b}$ bones, and the initial Jacobian matrices $\{\jac_i^0\}$ are the spatial gradients of the skinning weight field at the corresponding initial states:
\begin{align}
\point^{0}_i = \bone_i^{-1} \cdot \point' \quad
\jac^{0}_i = \pardev{\mathbf{d}_{\sigma_w}(\point, \bone)}{\point}\bigg|_{\point = \point_i^0} 
\label{equ:init}
\end{align}
The final set of correspondences is determined by their convergence:
\begin{align}
\mathcal{X}^* = \left\{\point^*_i \mid \left \| \mathbf{d}_{\sigma_w}(\point^*_i, \bone) - \point' \right \|_2 < \epsilon \right\} ,
\label{equ:root_set}
\end{align}
where $\epsilon$ is the convergence threshold. 

\boldparagraph{Aggregating Multiple Correspondences}
The maximum over the occupancy probabilities of all canonical correspondences gives the final occupancy prediction: 
\begin{align}
\occ'(\point',\mathbf{p}) = \max_{\point^* \in \mathcal{X}^*} \{  \function{f}(\point^*,\mathbf{p}) \} .
\label{equ:compose}
\end{align}

\boldparagraph{Losses} The canonical neural fields and the skinning weights can be learned jointly from observations in the deformed space. SNARF assumes direct 3D supervision and uses the binary cross entropy loss $\loss_{BCE}(\occ(\point',\mathbf{p}), \occ_{gt}(\point'))$ between the predicted and ground-truth occupancy for any deformed point. In addition, two auxiliary losses are applied during the first epoch to bootstrap training. SNARF randomly samples points along the bones that connect joints in canonical space and encourages their occupancy probabilities to be one. Moreover, SNARF encourages the skinning weights of all joints to be $1$ for their parent bones.

\boldparagraph{Gradients}
To learn the skinning weights $\mfunction{w}$ using a loss applied on the predicted occupancy probability in posed space $\loss(\occ(\point',\mathbf{p}))$, the gradient of $\loss$ \wrt \ $\sigma_w$ is required. Applying the chain rule, the gradient $\pardev{\loss}{\sigma_w}$ is given by
\begin{align}
\pardev{\loss}{\sigma_w} = \pardev{\loss}{\occ} \cdot \pardev{\occ}{\function{f}} \cdot \pardev{\function{f}(\point^*)}{\point^*} \cdot \pardev{\point^*}{\sigma_w},
\label{equ:grad}
\end{align}
where $\point^*$ is the root as defined in \eqnref{equ:root_set} 

The last term cannot be obtained using standard auto-differentiation because $\point^*$ is determined by the iterative correspondence search using ${\sigma_w}$. This iterative procedure is not trivially differentiable. To overcome this problem, implicit differentiation is used to derive the following analytical form of the last term:
\begin{align}
\pardev{\point^*}{\sigma_w} = - \left(\pardev{\mathbf{d}_{\sigma_w}(\point^*,\bone)}{\point^*} \right)^{-1}\cdot \pardev{ \mathbf{d}_{\sigma_w}(\point^*,\bone)}{\sigma_w} . 
\label{equ:def_grad}
\end{align}
Substituting \eqnref{equ:def_grad} into \eqnref{equ:grad} yields the gradient term $\pardev{\loss}{\sigma_w}$ which then allows skinning weights to be learned with standard back-propagation.

\section{Fast Differentiable Forward Skinning}
\label{sec:fast_snarf}

While the formulation mentioned above can articulate neural fields with good quality and generalization ability, the original SNARF algorithm is computationally expensive, which limits its wider application. As a reference, determining the correspondences of 200k points takes 800ms on an NVIDIA Quadro RTX 6000 GPU. In the following, we describe how \name \ overcomes this issue, reducing the computation time from 800ms to 5ms (\tabref{tab:ablation}).

\subsection{Voxel-based Correspondence Search}
The core of our method is to factor out costly computations at each root finding iteration in SNARF, including MLP evaluations and LBS calculations, into a pre-computation stage as illustrated in ~\algoref{algo}.

\boldparagraph{Voxel-based Skinning Field} The main speed bottleneck of SNARF lies in computing \eqnref{equ:broyden} at each iteration of Broyden's method. Computing \eqnref{equ:broyden} is time-consuming because it involves querying skinning weights, which are parameterized via an MLP in SNARF, and then computing LBS. We notice that the skinning weight field does not contain high-frequency details as illustrated in \figref{fig:skinning}. Therefore, we re-parameterize the skinning weight field $\mathbf{w}$ with a low-resolution voxel grid $\{\mathbf{w}_v\}$ with skinning weights $\mathbf{w}_v$ defined for each grid point $\point_v$. The skinning weights of any, non-grid aligned point in space are then obtained via tri-linear interpolation. We found that a resolution of $64\times64\times16$ is sufficient to describe the skinning weights in all experiments. Note that we use lower resolution along the $z$-axis due to the ``flatness" of the human body along this dimension in canonical space.

\boldparagraph{Pre-computing LBS} 
Computing linear blend skinning (\eqnref{equ:lbs}) at each root finding iteration also impacts  speed.
To further improve computation efficiency, we note that an explicit voxel-based skinning weights representation $\{\mathbf{w}_v\}$ allows us to compute the linearly blended skinning transformations for grid points $\{\mathbf{T}_v\}$ given current body poses:
\begin{align}
\mathbf{T}_v = \sum_{i=1}^{n_{\text{b}}}w_{v,i} \cdot \bone_i .
\end{align}
Then, during root finding, the required transformation at any canonical point $\mathbf{T}(\point)$ can be determined by tri-linearly interpolating neighbouring transformations in $\{\mathbf{T}_v\}$. Thus, LBS only needs to be run for a small set of grid points instead of all query points in the root finding procedure. 

\begin{figure}[t!]
    \centering
    \includegraphics[width=\linewidth,trim=0 10 0 0,clip]{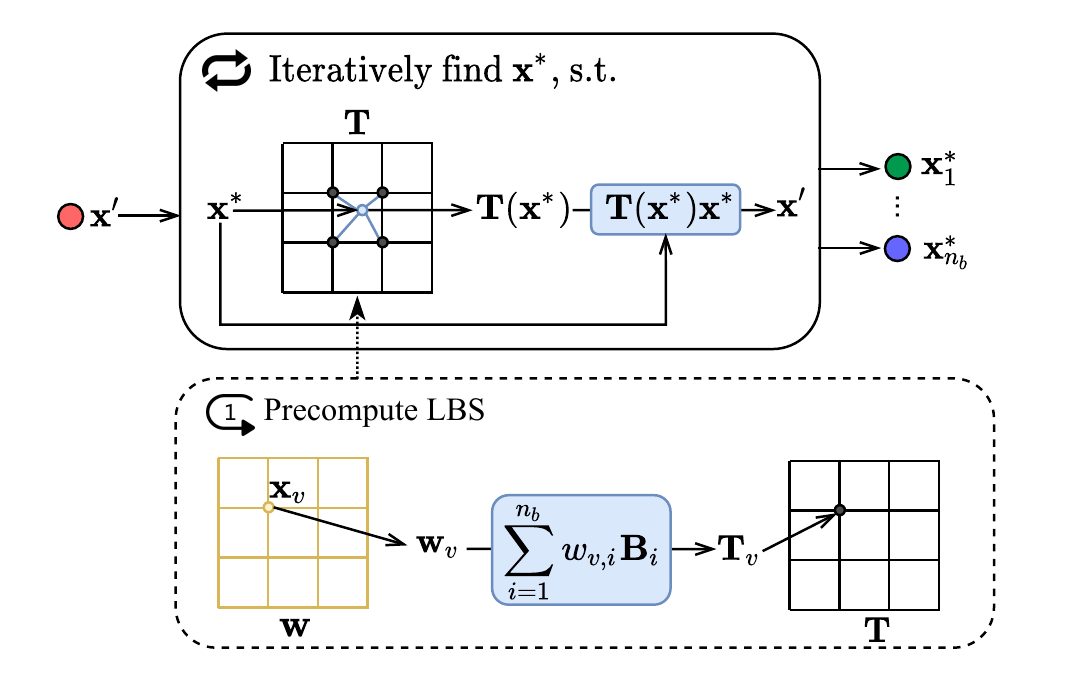}
    \caption{\textbf{Voxel-based Forward Skinning~(\name).} In comparison with SNARF (cf. \figref{fig:snarf}), \name \ uses a voxel-based representation to speed up the iterative correspondence search. The skinning weight field is represented as a voxel grid. For each pose, we first pre-compute LBS for each grid point, yielding a transformation field. For each query deformed point $\point'$, \name \ finds its canonical correspondences $\point^*$ which satisfy $\mathbf{T}(\point^*)\cdot\point^*=\point'$.
    \vspace{-1em}
    }
    \label{fig:fast_snarf}
\end{figure}

\boldparagraph{Custom CUDA Kernel} Broyden's method is iterative and involves many small operations that have to be computed per query point, such as  arithmetic operations on small matrices and reading values from the voxel grid. We note that these operations can be computed in an independent manner. This motivates us to implement this module with a custom CUDA kernel instead of using native functions in standard deep learning frameworks. The handwritten kernel, parallelized over query points, fuses the entire method into a single kernel that keeps working variables in registers, avoiding unnecessary time and memory costs from launching native kernels and synchronizing intermediate results. The input to our CUDA kernel for iterative root finding is the pre-computed voxel grid of transformations $\{\mathbf{T}_{v}\}$, the bone transformations $\mathbf{B}$ as well as query points $\point'$. The kernel first computes the multiple initialization states (\eqnref{equ:init}). Then, at each root finding iteration, the kernel tri-linearly interpolates $\{\mathbf{T}_{v}\}$ and transforms the points (\eqnref{equ:lbs2}), and applies Broyden's update (\eqnref{equ:broyden}). After each iteration $k$, we filter diverged and converged points $\point^k$ by checking whether $\left \| \mathbf{d}_{\sigma_w}(\point^k, \bone) - \point' \right \|_2$  is larger than the divergence threshold or smaller than the convergence threshold, further reducing the number of required computations. 
\newcommand\Algphase[1]{%
\vspace*{-.7\baselineskip}\Statex\hspace*{\dimexpr-\algorithmicindent+1em\relax}\rule{\linewidth}{0.4pt}%
\Statex\hspace*{-0.5em}\textbf{#1}%
\vspace*{-.7\baselineskip}\Statex\hspace*{\dimexpr-\algorithmicindent+1em\relax}\rule{\linewidth}{0.4pt}%
}

\begin{algorithm}[t]
\setstretch{1.2}
\centering
\caption{Correspondence Search}
\label{algo}
\begin{algorithmic}[0]
\State \textbf{Inputs:}
\Statex $\{ (\mathbf{x}',\point^0,\tilde{\jac}^0)\}$ query points and initialization
\Statex $\bone$ bone transformations
\Statex $\mathbf{w}_{\sigma_w}$ skinning weights MLP
\Algphase{Variant 1: MLP-based Search (SNARF)}
\For {$\point',\point^0,\tilde{\jac}^0 \in \{ (\mathbf{x}',\point^0,\tilde{\jac}^0)\}$ \textbf{in parallel}}
        \For  {$k \gets 0,n $}
            \setulcolor{red}
            \State \ul{$w_1, ..., w_{n_b} \gets \mathbf{w}_{\sigma_w}(\mathbf{x}^k_j)$} \hfill {\color{red}{costly operations}}
            \State \ul{$\mathbf{T} \gets \sum_{i=1}^{n_{\text{b}}}{w}_{i}(\point^k) \cdot \bone_i$}\hfill{\color{red}{inside root finding}}
            \State $\point^{k+1}, \tilde{\jac}^{k+1} \gets \text{broyden}(\point^{k}, \tilde{\jac}^{k},\mathbf{T},\point')$\Comment{\eqnref{equ:root_equ}}
        \EndFor
\EndFor
\State \textbf{return:} $\{\mathbf{x}^n\}$
\Algphase{Variant 2: Voxel-based Search (\name)}

\For {\textbf{each} $\point_v \in \{\point_v\}$ \textbf{in parallel}}
    \setulcolor{blue}
    \State \ul{$w_1, ..., w_{n_b} \gets \mathbf{w}_{\sigma_w}(\mathbf{x}_v)$}\hfill{\color{blue}{pre-computation}}
    \State \ul{$\mathbf{T}_v \gets \sum_{i=1}^{n_{\text{b}}}{w}_{i} \cdot \bone_i$}
\EndFor
\For {$\point',\point^0,\tilde{\jac}^0 \in \{ (\mathbf{x}',\point^0,\tilde{\jac}^0)\}$ \textbf{in parallel}} 
    \For {$k \gets 0,n $}
        \State \ul{$\mathbf{T} \gets \text{trilerp}(\point^k,\{\mathbf{T}_v\})$}\hfill{\color{blue}{lightweight operation}}
        \State $\point^{k+1}, \tilde{\jac}^{k+1} \gets \text{broyden}(\point^{k}, \tilde{\jac}^{k},\mathbf{T},\point')$ \Comment{\eqnref{equ:root_equ}}
    \EndFor
\EndFor
\State \textbf{return:} $\{\mathbf{x}^n\}$
\end{algorithmic}
\end{algorithm}

\boldparagraph{Remove Duplicate Correspondences} A further important speed optimization pertains to the treatment of multiple correspondences found by the root finding algorithm. The set of valid correspondences contains duplicates because different initial states can converge to the same solution. To avoid unnecessary evaluation of the canonical neural fields for these duplicates, we detect duplicate solutions by their relative distances in canonical space and discard them.

\subsection{Skinning Weights Optimization}

Analogous to SNARF, in theory, \name \ supports learning skinning weights with the analytical gradients in  \eqnref{equ:def_grad}. However, there are two practical challenges. 

\boldparagraph{Approximated Gradient} A first problem lies in that \eqnref{equ:def_grad} involves computing derivatives and the matrix inversion $\left(\pardev{\mathbf{d}_{\sigma_w}(\point^*,\bone)}{\point^*} \right)^{-1}$, which is time-consuming, impeding our goal of fast training. To address this, we note that this term is identical to the inverse of the Jacobian $\jac$ in the last iteration of root finding (\eqnref{equ:broyden}):
\begin{align}
\left(\pardev{\mathbf{d}_{\sigma_w}(\point^*,\bone)}{\point^*} \right)^{-1} = {\underbrace{\left(\pardev{\mathbf{d}_{\sigma_w}(\point^*,\bone)-\point'}{\point^*} \right)}_{\jac}}^{-1} 
\label{equ:jac_grad}
\end{align}
because the deformed point $\point'$ is a given input and is independent of the canonical correspondence $\point^*$. The inverse of the Jacobian $\jac$ is approximated in Broyden's method as $\Tilde{\jac}$. Thus, we use $\Tilde{\jac}$ directly:
\begin{align}
\pardev{\point^*}{\sigma_w} = - \Tilde{\jac}\cdot \pardev{ \mathbf{d}_{\sigma_w}(\point^*,\bone)}{\sigma_w} . 
\label{equ:jac_grad}
\end{align}

\boldparagraph{Distilling Smooth Skinning Fields} 
A second problem is that the voxel-based parameterization does not have the global smoothness bias of MLPs, thus optimizing voxels directly would result in a noisy skinning weight field. To obtain smooth skinning weights while using voxel-based correspondence search, a common approach is to apply a total variational regularizer. However, we experimentally found that this regularization does not lead to the desired smoothness of the skinning weights and negatively affects the accuracy of the generated shapes. We thus propose a new approach by using an MLP to parameterize the skinning weight field during training but continuously distill the MLP to a voxel-based skinning weight field at each training iteration. The skinning weight field is thus smooth by design due to the intermediate use of an MLP. At each training iteration, we compute the skinning weights voxel grid on the fly by evaluating the MLP at grid points $\{\point_v\}$, and then use our fast voxel-based correspondence search. In this scheme the parameters of the MLP are optimized during training, not the voxels directly which are only used to store the weights. The conversion from MLP to voxels does introduce additional computation during training, but the overhead is minor since the voxel grid is low resolution. The inference speed is not influenced at all because the MLP is used during training only. This yields on-par accuracy with SNARF as we inherit the inductive smoothness bias of the MLP-based skinning weight model. 

\subsection{Learning Avatars from 3D Scans}
We can use our articulation module to learn animatable human avatars with realistic cloth deformations from 3D scans. Given a set of 3D meshes in various body poses, our method learns the human shape in canonical space as an occupancy field alongside the canonical skinning weight field which is needed for animation.
We model the canonical shape using an occupancy field and use the same training losses as \snarf~\cite{Chen2021ICCV} (see \secref{sec:snarf}).

\section{Experiments}
\label{sec:exp}
\subsection{Minimally Clothed Humans}
\label{sec:exp_minimal}
We first evaluate the speed and accuracy of our method and baselines on minimally clothed humans.

\subsubsection{Dataset} 
We follow the same evaluation protocol as NASA~\cite{Deng2020ECCV} and SNARF~\cite{Chen2021ICCV}. More specifically, we use the DFaust~\cite{Bogo2017CVPR} subset of AMASS~\cite{Mahmood2019ICCV} for training and evaluating our model on SMPL meshes of people in minimal clothing. This dataset covers 10 subjects of varying body shapes. For each subject, we use 10 sequences, from which we randomly select one sequence for validation, using the rest for training. For each frame in a sequence, 20K points are sampled, among which, half are sampled uniformly in space and half are sampled in near-surface regions by first applying Poisson disk sampling on the mesh surface, followed by adding isotropic Gaussian noise with $\sigma=0.01$ to the sampled point locations. 
In addition to the ``within distribution'' evaluation on DFaust, we test ``out of distribution" performance on another subset of AMASS, namely PosePrior~\cite{Akhter2015CVPR}.
This subset contains challenging, extreme poses, not present in DFaust.

\begin{table*}[t!]
\centering
\resizebox{0.9\textwidth}{!}{
\begin{tabular}{lcccccccc}
\toprule
\multicolumn{1}{l|}{} & \multicolumn{2}{c|}{Within Distribution}                      & \multicolumn{2}{c|}{Out of Distribution}                      & \multicolumn{3}{c|}{Inference Speed}   & \multicolumn{1}{c}{Training Time}                                    \\ \cline{2-8} 
\multicolumn{1}{l|}{} & \multicolumn{1}{c|}{IoU bbox} & \multicolumn{1}{c|}{IoU surf} & \multicolumn{1}{c|}{IoU bbox} & \multicolumn{1}{c|}{IoU surf} & \multicolumn{1}{c}{Articulation} & \multicolumn{1}{c}{Shape} & \multicolumn{1}{c|}{Total} \\ \midrule
Pose-ONet*             & 79.34\%                               & 58.61\%                              & 49.21\%                              & 28.69\%                              &   0ms                           &  28.95ms              & 29.88ms       &    16min    \\
Backward-LBS*              & 81.68\%                              & 87.44\%                             & 66.93\%                             & 68.93\%                              &   12.39ms                           &  27.67ms              &  40.60ms    &  31min       \\
\hline
NASA                  & 96.14\%                             & 86.98\%                              & 83.16\%        & 60.21\%                              &   -                           &    -          &  582ms       &    4h     \\
SNARF                 & 97.31\%                              & 90.38\%                              & 93.97\%                              & 80.65\%                              & 806.67ms                     & 186.82ms        &  994.01ms                  &       8h    \\
\name             &   \textbf{97.41\% }                           &   \textbf{90.52\%}                           &   \textbf{94.20\%}                            &   \textbf{81.25\%}                            & 5.27ms                              &   27.78ms                  & 34.70ms         & 25min \\
\bottomrule
\end{tabular}
}
\vspace{-0.5em}
\caption{\textbf{Quantitative Results on Minimally Clothed Humans.} The mean IoU of uniformly sampled points in space (IoU bbox) and points near the surface (IoU surface), as well as the inference and training time are reported. Our method achieves similar accuracy as SNARF (previous state-of-the-art) while being much faster. Our method outperforms all other baselines in terms of accuracy. Improvements are more pronounced for points near the surface, and for poses outside the training distribution. Also our method is faster than all baselines except Pose-ONet. Note that Pose-ONet and Backward-LBS (above the separation line, marked with *) produce distorted shapes, as shown in \figref{fig:dfaust}.
}
\vspace{-1em}
\label{tab:dfaust}
\end{table*}

\begin{figure*}[t!]
\begin{center}
\setlength\tabcolsep{1pt}
\newcommand{\crop}{0.8cm}
\newcommand{\cropsmall}{0.4cm}
\newcommand{\height}{2.55cm}
\begin{tabularx}{\linewidth}{ l cc| ccccc }
\rotatebox{90} {~~~~~Pose-ONet~~~~~}
                        &   \includegraphics[height=\height, trim={-1cm 0 {\cropsmall} 0},clip]{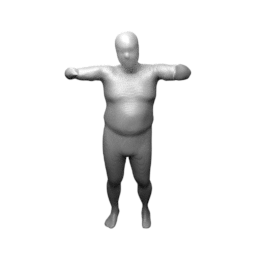}
                        &   \includegraphics[height=\height, trim={{\crop} 0 {\crop} 0},clip]{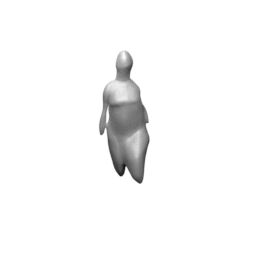}
                        &\includegraphics[height=\height, trim={-1cm 0 {\cropsmall} 0},clip]{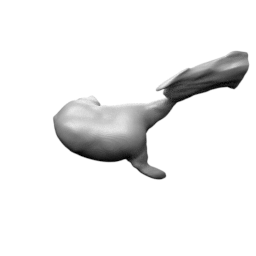}
                        &  \includegraphics[height=\height, trim={{\cropsmall} 0 {\cropsmall} 0},clip]{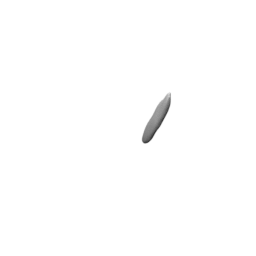}
                        &   \includegraphics[height=\height, trim={{\cropsmall} 0 {\cropsmall} 0},clip]{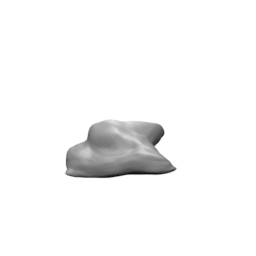}
                        &   
                        &  \includegraphics[height=\height, trim={{\cropsmall} 0 {\cropsmall} 0},clip]{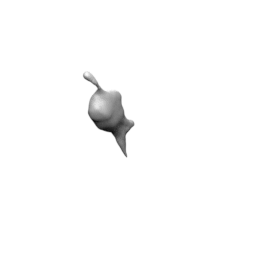}\\
\rotatebox{90} {~~~~~Backward~~~~~} 
                        &   \includegraphics[height=\height, trim={-1cm 0 {\cropsmall} 0},clip]{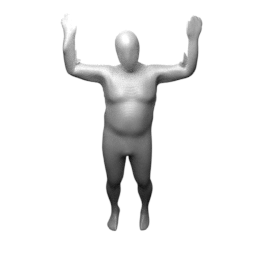}
                        &   \includegraphics[height=\height, trim={{\crop} 0 {\crop} 0},clip]{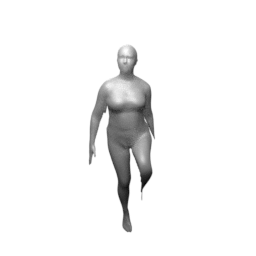}
                        &\includegraphics[height=\height, trim={-1cm 0 {\cropsmall} 0},clip]{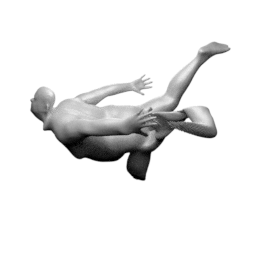}
                        &   \includegraphics[height=\height, trim={{\cropsmall} 0 {\cropsmall} 0},clip]{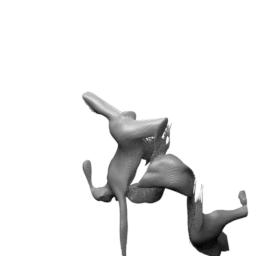}
                        &  \includegraphics[height=\height, trim={{\cropsmall} 0 {\cropsmall} 0},clip]{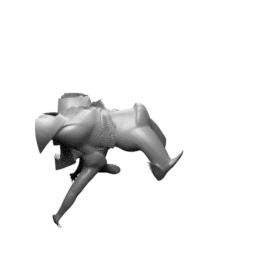}
                        &   \includegraphics[height=\height, trim={{\cropsmall} 0 {\cropsmall} 0},clip]{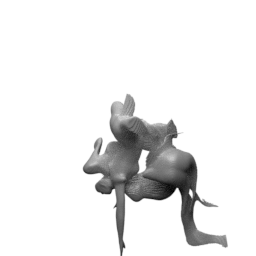}
                        &   \includegraphics[height=\height, trim={{\cropsmall} 0 {\cropsmall} 0},clip]{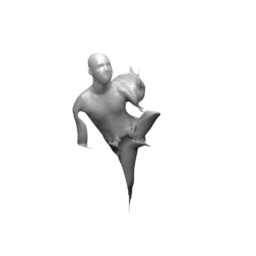}\\
\rotatebox{90} {~~~~~~~NASA~~~~~~}
                        &   \includegraphics[height=\height, trim={-1cm 0 {\cropsmall} 0},clip]{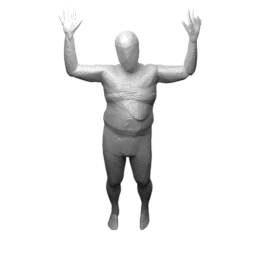}
                        &   \includegraphics[height=\height, trim={{\crop} 0 {\crop} 0},clip]{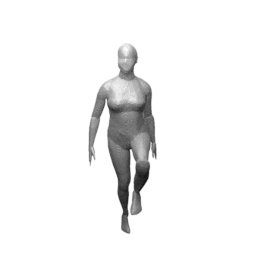}
                        &\includegraphics[height=\height, trim={-1cm 0 {\cropsmall} 0},clip]{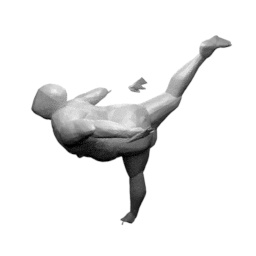}
                        &   \includegraphics[height=\height, trim={{\cropsmall} 0 {\cropsmall} 0},clip]{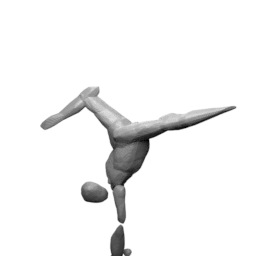}
                        &   \includegraphics[height=\height, trim={{\cropsmall} 0 {\cropsmall} 0},clip]{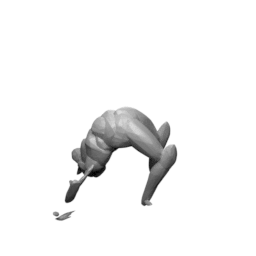}
                        &   \includegraphics[height=\height, trim={{\cropsmall} 0 {\cropsmall} 0},clip]{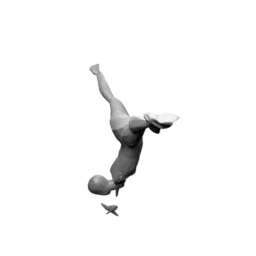}
                        &   \includegraphics[height=\height, trim={{\cropsmall} 0 {\cropsmall} 0},clip]{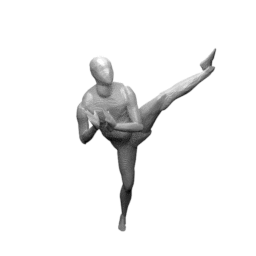}\\
\rotatebox{90} {~~~~~~~~~~SNARF~~~~~~~~}
                        &   \includegraphics[height=\height, trim={-1cm 0 {\cropsmall} 0},clip]{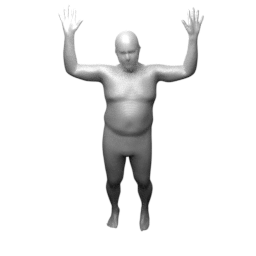}
                        &   \includegraphics[height=\height, trim={{\crop} 0 {\crop} 0},clip]{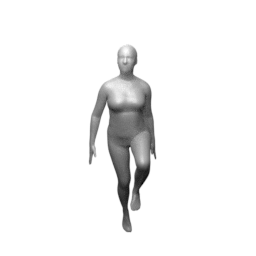}
                        &\includegraphics[height=\height, trim={-1cm 0 {\cropsmall} 0},clip]{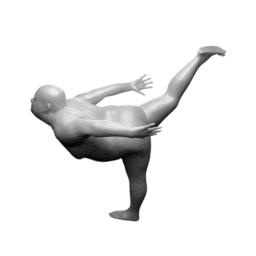}
                        &   \includegraphics[height=\height, trim={{\cropsmall} 0 {\cropsmall} 0},clip]{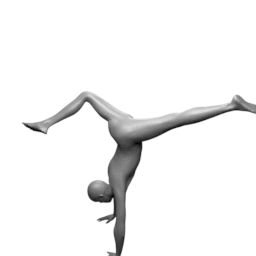}
                        &   \includegraphics[height=\height, trim={{\cropsmall} 0 {\cropsmall} 0},clip]{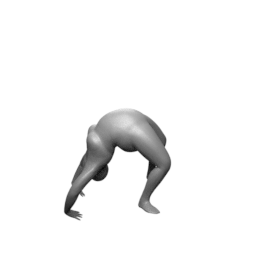}
                        &   \includegraphics[height=\height, trim={{\cropsmall} 0 {\cropsmall} 0},clip]{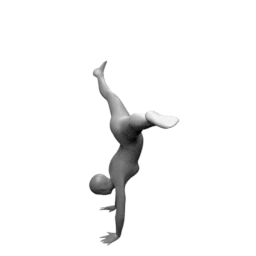}
                        &   \includegraphics[height=\height, trim={{\cropsmall} 0 {\cropsmall} 0},clip]{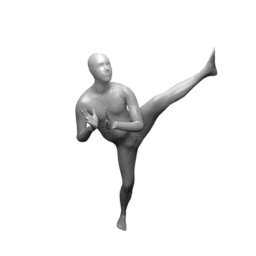}\\
\rotatebox{90} {~~~~~~~Fast-SNARF~~~~~~}
                        &   \includegraphics[height=\height, trim={-1cm 0 {\cropsmall} 0},clip]{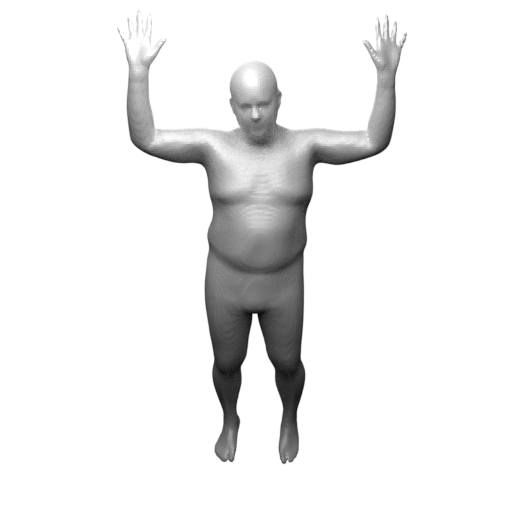}
                        &   \includegraphics[height=\height, trim={{\crop} 0 {\crop} 0},clip]{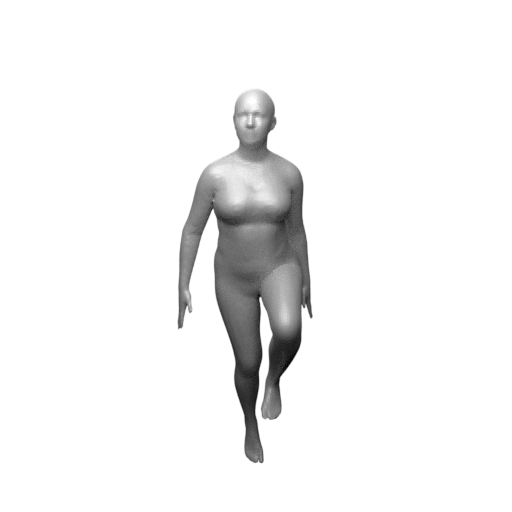}
                        &\includegraphics[height=\height, trim={-1cm 0 {\cropsmall} 0},clip]{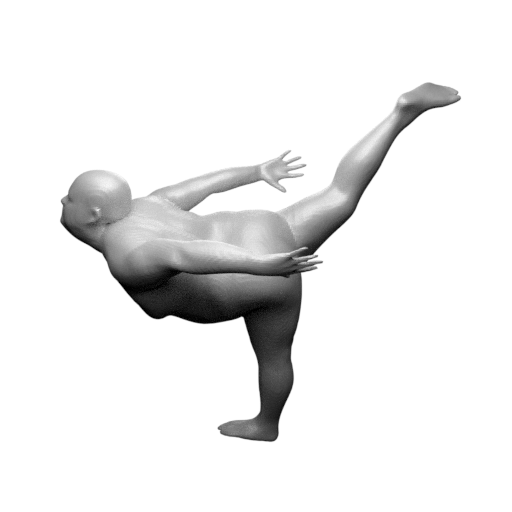}
                        &   \includegraphics[height=\height, trim={{\cropsmall} 0 {\cropsmall} 0},clip]{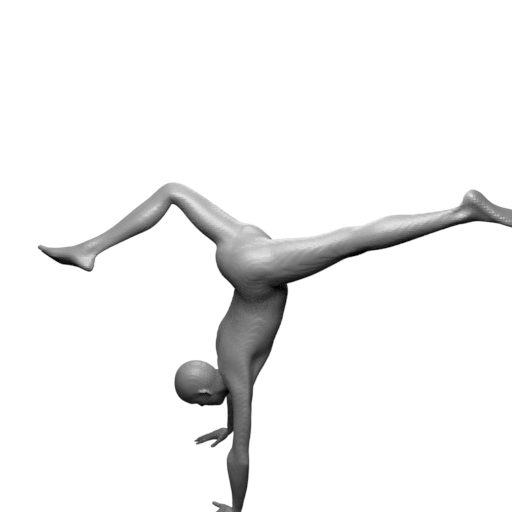}
                        &   \includegraphics[height=\height, trim={{\cropsmall} 0 {\cropsmall} 0},clip]{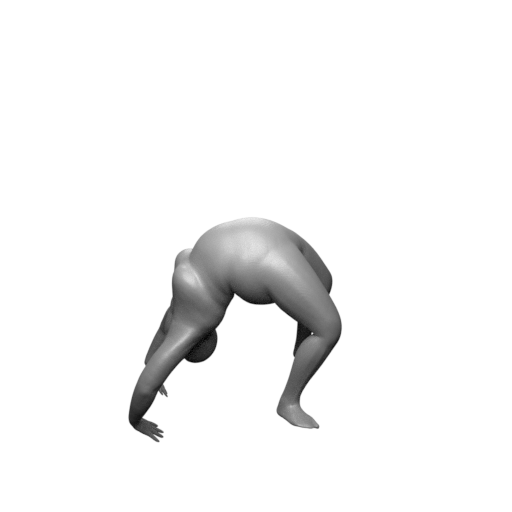}
                        &   \includegraphics[height=\height, trim={{\cropsmall} 0 {\cropsmall} 0},clip]{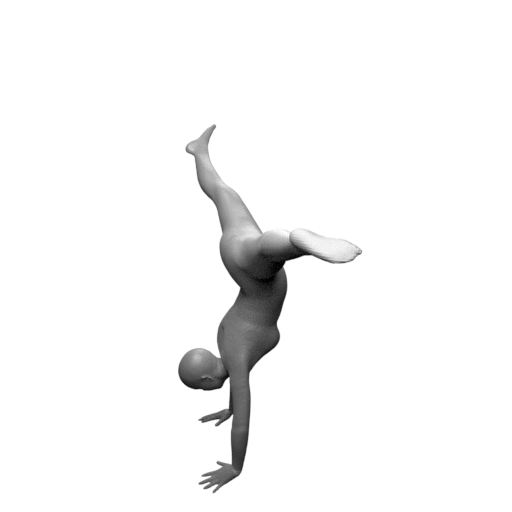}
                        &   \includegraphics[height=\height, trim={{\cropsmall} 0 {\cropsmall} 0},clip]{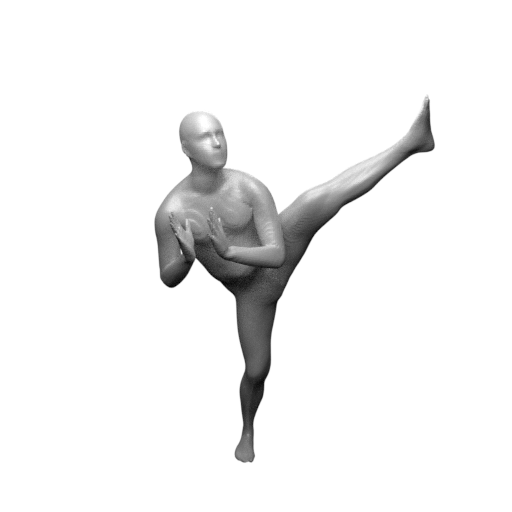}\\
\rotatebox{90} {~~~Ground Truth~~~}  
                        &   \includegraphics[height=\height, trim={-1cm 0 {\cropsmall} 0},clip]{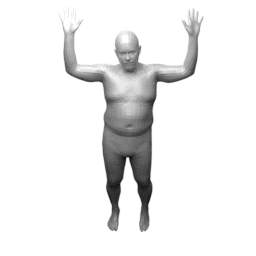}
                        &   \includegraphics[height=\height, trim={{\crop} 0 {\crop} 0},clip]{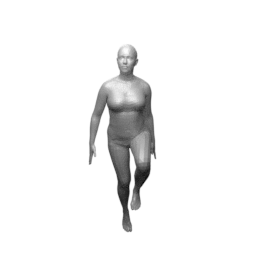}
                        &\includegraphics[height=\height, trim={-1cm 0 {\cropsmall} 0},clip]{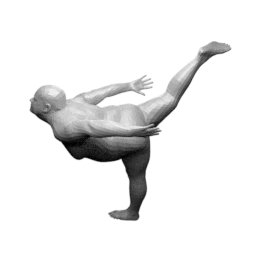}
                        &   \includegraphics[height=\height, trim={{\cropsmall} 0 {\cropsmall} 0},clip]{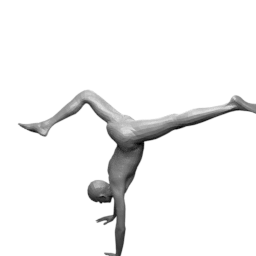}
                        &   \includegraphics[height=\height, trim={{\cropsmall} 0 {\cropsmall} 0},clip]{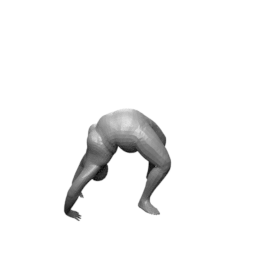}
                        &   \includegraphics[height=\height, trim={{\cropsmall} 0 {\cropsmall} 0},clip]{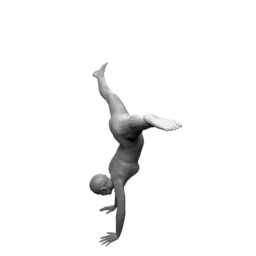}
                        &   \includegraphics[height=\height, trim={{\cropsmall} 0 {\cropsmall} 0},clip]{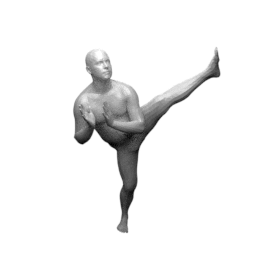}\\
                        &\multicolumn{2}{c}{\centering Within Distribution} &\multicolumn{5}{c}{\centering Out of Distribution}\\

\end{tabularx}
\vspace{-0.5em}
\caption{\textbf{Qualitative Results on Minimally Clothed Humans.} 
Our method and SNARF produce results similar to the ground-truth with correct pose and plausible local details, both for poses within the training distribution and more extreme (OOD) poses. In contrast, the baseline methods suffer from various artifacts including incorrect poses (Pose-ONet), degenerate shapes (Pose-ONet and Backward), and discontinuities near joints (NASA), which become more severe for unseen poses.
}
\vspace{-1.5em}
\label{fig:dfaust}
\end{center}
\end{figure*}

\subsubsection{Baselines}
\noindent We consider SNARF as our main baseline. In addition, we consider the following baselines. For SNARF, ``Back-LBS" and ``Pose-ONet" we use the same training losses and hyperparameters as in \name.

\boldparagraph{Pose-Conditioned Occupancy Networks (Pose-ONet)}
This baseline extends Occupancy Networks~\cite{Mescheder2019CVPR} by directly concatenating the pose input to the occupancy network.

\boldparagraph{Backward Skinning {(Back-LBS)}} 
This baseline implements the concept of backward skinning similar to~\cite{Jeruzalski2020ARXIV}. A network takes a deformed point and pose condition
as input and outputs the skinning weights of the deformed point. The deformed point is then warped back to canonical space via LBS and the canonical correspondence is fed into the canonical shape network to query occupancy.

\boldparagraph{NASA}
NASA~\cite{Deng2020ECCV} models articulated human bodies as a composition of multiple parts, each of which transforms rigidly and deforms according to the pose. Note that in contrast to us, NASA requires ground-truth skinning weights for surface points as supervision. 
We use the official NASA implementation provided by the authors.

\subsubsection{Results and Discussion}

\boldparagraph{Within Distribution Accuracy}
Overall, all methods perform well in this relatively simple setting, as shown in \tabref{tab:dfaust}. Our method achieves on-par or better accuracy compared to SNARF and provides an improvement over other baselines. 
Our method produces bodies with smooth surfaces and correct poses as shown in \figref{fig:dfaust}. In contrast, NASA suffers from discontinuous artifacts near joints. Back-LBS and Pose-ONet suffer from missing body parts.

\boldparagraph{Out of Distribution (OOD) Accuracy}
In this setting, we test the trained models on a different dataset, PosePrior~\cite{Akhter2015CVPR}, to assess the performance in more realistic settings, where poses can be far from those in the training set. Unseen poses cause drastic performance degradation to the baseline methods as shown in~\tabref{tab:dfaust}. In contrast, similar to SNARF, our method degrades gracefully despite test poses being drastically different from training poses and very challenging. As can be seen in~\figref{fig:dfaust}, our method generates natural shapes for the given poses while NASA fails to generate correctives at bone intersections for unseen poses, leading to noticeable artifacts. Pose-ONet fails to generate meaningful shapes and Back-LBS produces distorted bodies due to incorrect skinning weights.

\boldparagraph{Speed Comparison} We report the training and inference speed of all methods on a single NVIDIA Quadro RTX 6000 GPU. In this setting, with MLP-based canonical shape, \name \ can be trained within 25 minutes and produces accurate shapes in any pose. Baseline methods that reach similar speed, \ie \ Pose-ONet, and Back-LBS, do not produce satisfactory results (see \figref{fig:dfaust}). Compared to the original SNARF, our improvements, detailed in \secref{sec:fast_snarf}, lead to a speed-up of $150\times$ for the articulation module without loss of accuracy, as shown in \tabref{tab:dfaust}. \name \ also dramatically boosts the training speed (25 minutes \vs~8 hours). Compared to NASA, \name \ evaluates the canonical shape MLP only for true correspondences, while NASA always generates many candidate correspondences, one for each bone, and needs to evaluate the canonical shape MLP for all candidates, leading to slow inference (582ms vs.~35ms) and training (4 hours \vs~25 minutes).

\begin{figure}[t]
\begin{center}
\setlength\tabcolsep{1pt}
\begin{tabularx}{\textwidth}{ccc}
SNARF & \name & w/o MLP distillation \\
8 h & 25 min& 23 min \\

    \includegraphics[width=0.3\linewidth,trim=0 512 512 1024, clip]{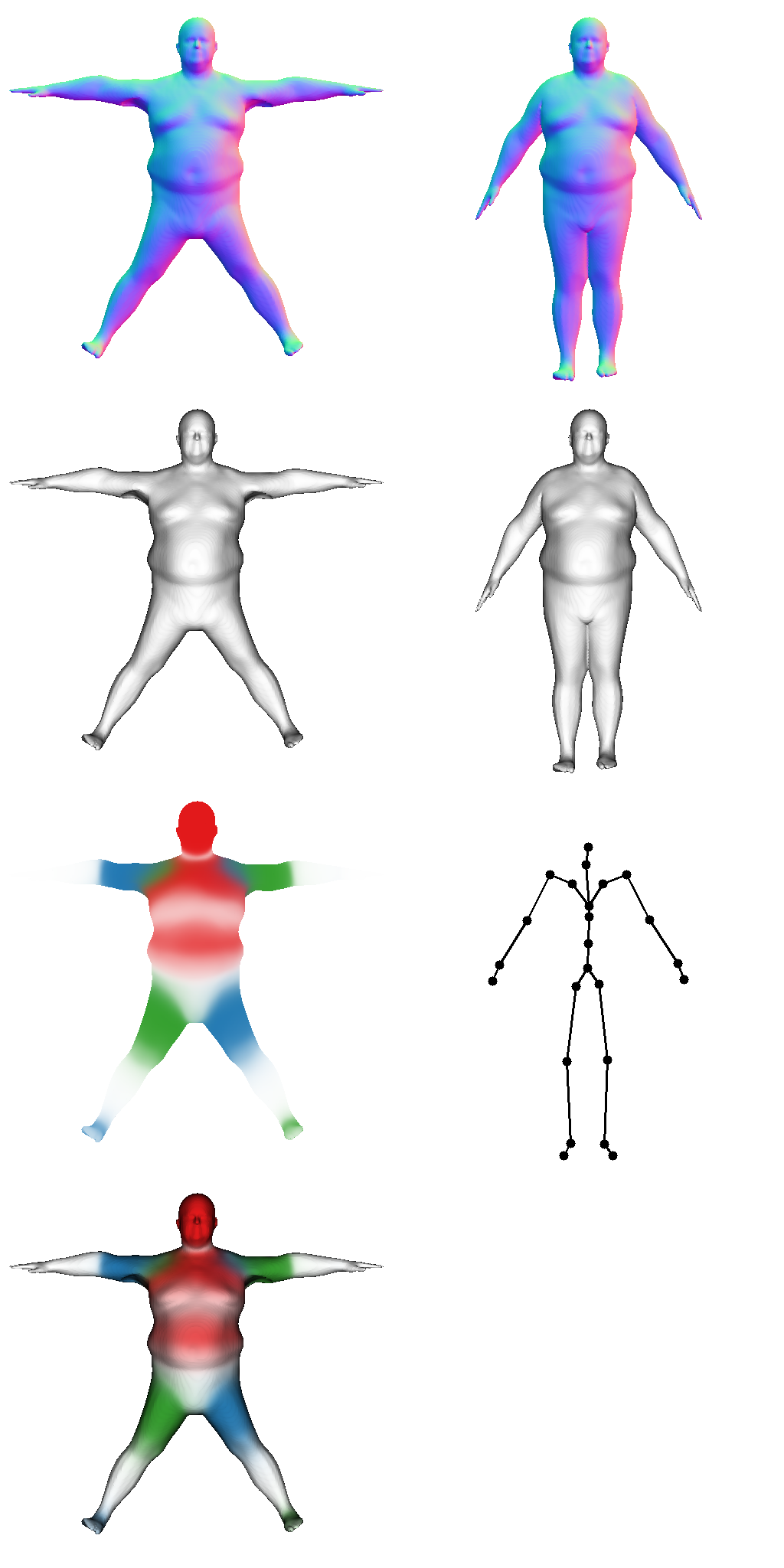}    
&   \includegraphics[width=0.3\linewidth,trim=0 512 512 1024, clip]{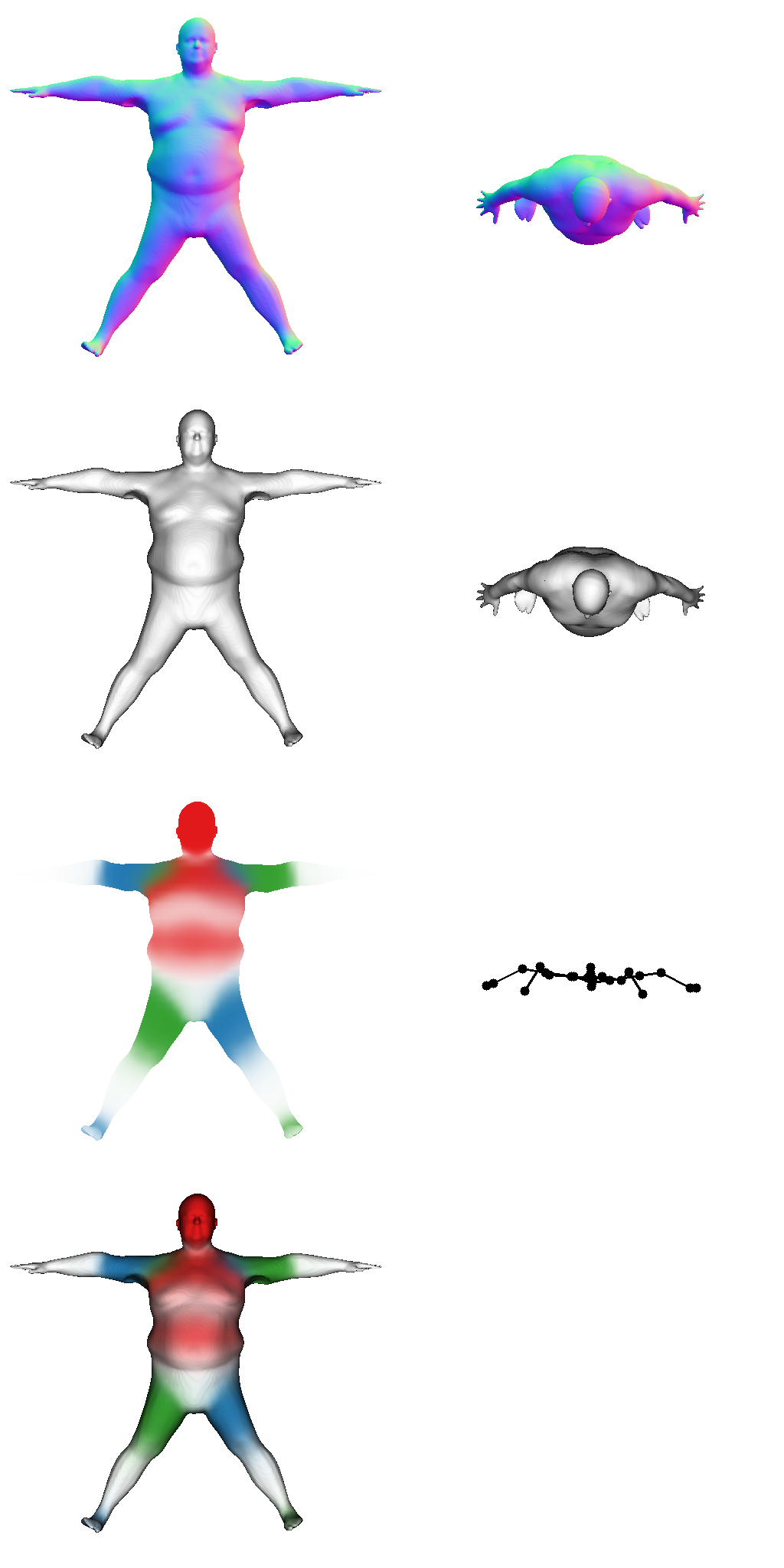}
&   \includegraphics[width=0.3\linewidth,trim=0 512 512 1024, clip]{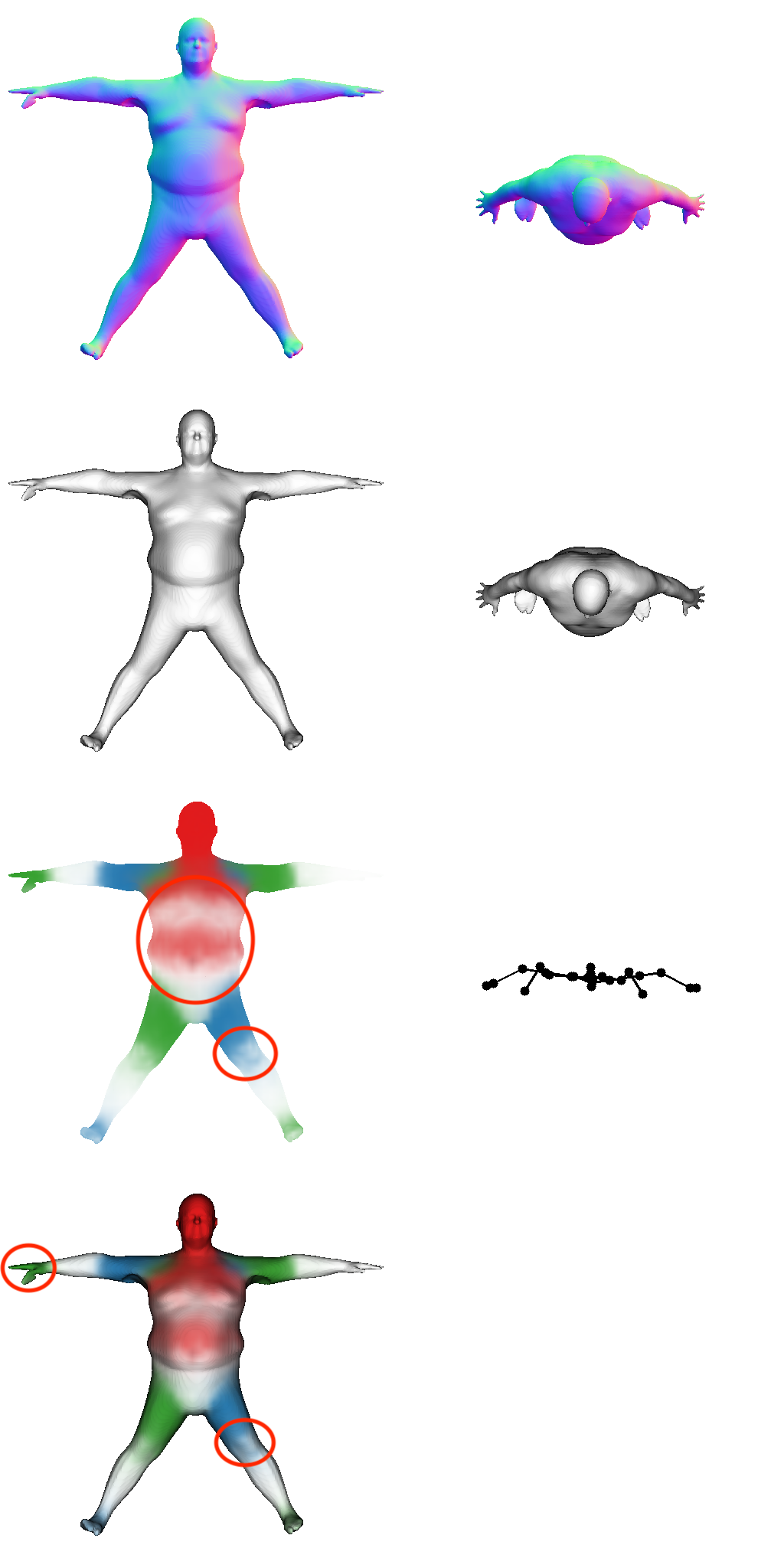} \\
\end{tabularx}
\vspace{-0.5em}
\caption{\textbf{Skinning Weight Learning Strategies.} We show skinning weights learned with three different strategies as well as the corresponding training times. See text.
}
\vspace{-1.5em}
\label{fig:skinning}
\end{center}
\end{figure}

\boldparagraph{Ablation - MLP Distillation}
SNARF optimizes an MLP-based skinning field, resulting in smooth skinning weights but slow training and inference. In \name, we adopt an MLP distillation strategy: we optimizes a MLP-based skinning weight field for smoothness, but convert it on the fly to a low resolution voxel grid at each training iteration, to enable voxel-based correspondence search. In this way, \name learns a similarly smooth skinning field as shown in \figref{fig:skinning}, yet is much faster than SNARF (see \tabref{tab:ablation}).

We also compare this MLP distillation strategy with a naive strategy in which we directly optimize the skinning weights at each grid point with an additional total variation loss on the skinning weights voxel grid. As shown in \tabref{tab:ablation}, directly optimizing skinning weights voxel (w/o MLP distillation) leads to inferior results. This accuracy degradation is due to noisy skinning weights as shown in \figref{fig:skinning}. In contrast, our strategy distills smooth skinning weights voxels from the MLP while introducing only a slight overhead during training (25 minutes \vs~23 minutes).

\boldparagraph{Ablation - Voxel Grid Resolution}
We study the effect of different resolutions of the skinning weight voxel grid. The results are shown in \tabref{tab:ablation}. In general, higher resolutions lead to higher accuracy but longer training and inference time. A resolution of $32\times32\times8$ or $64\times64\times16$ yields a good balance between accuracy and speed. A grid of lower resolution $16\times16\times4$ cannot fully represent the skinning weight field and leads to a noticeable accuracy degradation (by $2.8\%$). On the other hand, further increasing the resolution to $128\times128\times32$ produces diminishing returns, \ie \ only $0.3\%$ IoU improvement, because the skinning weight field is naturally smooth and does not contain high-frequency details. Also, higher resolution significantly slows down the training and inference speed by more than 2 times because 1) more points need to be evaluated when converting the MLP to voxels during training and 2) the high-resolution voxel grid does no longer fit into the GPU's shared memory and impacts read speeds significantly.

\begin{table}[t!]
\centering
\resizebox{\linewidth}{!}{
\begin{tabular}{@{}lccc@{}}
\toprule
Configurations & Accuracy & Inference & Training \\
\midrule
Baseline SNARF                      &  80.7\%    & 807ms + 187ms & 8h\\
+ Voxel-based search                &  -                & \ 61ms + 187ms &-\\
\quad + Pre-compute LBS             &  -                & \ 40ms + 187ms &-\\ 
\quad \quad + CUDA kernel           &  -                & \ \ 5.3ms + 187ms &-\\
\quad \quad \quad + Filter corres.  &  -                & \ \ 5.3ms + 28ms  &-\\
\midrule
\name \                          &  81.2\%              & \ \ 5.3ms + 28ms & 25 min\\
w/o MLP distillation                    &  78.2\%              & \ \ 5.3ms + 28ms & 23 min\\
\midrule
$16\times16\times4$ &78.3\% & 3.6ms + 28ms & 23min\\
$32\times32\times8$ &81.1\% & 4.6ms + 28ms& 24min \\
$64\times64\times16$ &81.2\% & 5.3ms + 28ms& 25min\\
$128\times128\times32$ &81.5\% & 16ms + 28ms& 52min\\

\bottomrule
\end{tabular}
}
\caption{
\textbf{Quantitative Ablation Study.} We report accuracy (the mean IoU of points near the surface in out of distribution setting), inference speed (articulation speed + shape query speed) and training time of several ablative baselines. 
}
\label{tab:ablation}
\end{table}

\begin{figure*}[t]
\includegraphics[width=\textwidth]{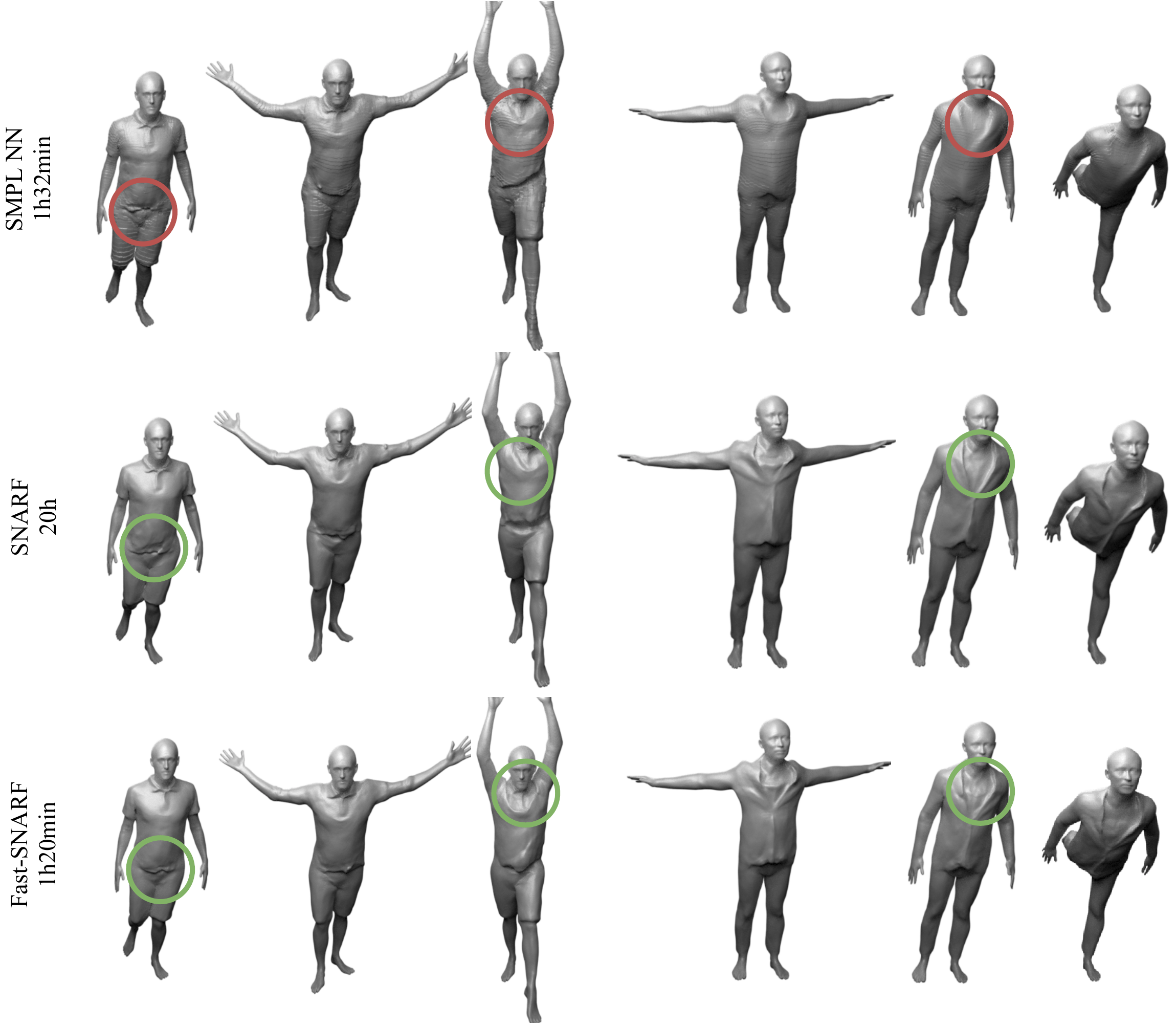}
\vspace{-2.5em}
\caption{\textbf{Qualitative Results on Clothed Humans~\cite{Ma2020CVPR}}.
Our method and SNARF both learn realistic clothing shape and deformations. In contrast, the baseline method using a skinned base mesh produces less details due to the inaccurate deformation when the base mesh mismatches the actual shape (highlighted in red circle).
\vspace{-1em}}
\label{fig:cape}
\end{figure*}

\subsection{Clothed Avatar from Scans}

\boldparagraph{Dataset} We use the registered meshes from CAPE~\cite{Ma2020CVPR} and their registered SMPL parameters to train our model. We use 8 subjects with different clothing types for evaluation.
We train a model for each subject and clothing condition.

\boldparagraph{Baselines}
Clothed humans are more challenging to model than minimally clothed humans due to the clothing details and non-linear deformations. Since most baselines from \secref{sec:exp_minimal} already suffer from implausible shapes and artifacts, we exclude them in this evaluation. Instead, we keep SNARF as our major baseline, and also include a new baseline denoted as ``SMPL NN". This baseline assumes that a skinned base mesh is given, such as SMPL~\cite{Loper2015SIGGRAPH}. Given a pose, such a method first deforms the SMPL model to the target pose using mesh-based LBS. Then for each query point in deformed space, its corresponding skinning weights are defined as the skinning weights of its nearest vertex on the deformed SMPL mesh. Finally, with the skinning weights, the query point can be transformed back to the canonical space base on inverse LBS.

\boldparagraph{Results} The results are shown in \figref{fig:cape}. Our method can generate realistic clothed humans in various poses including details on the face and clothing (\eg \ the collar on the left sample). The clothing also deforms naturally with the body poses (\eg \ the collar on the left sample and the lapel on the right sample). While SNARF produces results of similar quality, training our method only requires a fraction of SNARF's training time (80 minutes \vs \ 20 hours). Compared with the SMPL NN baseline, our results contain much more detail because our method derives accurate correspondences between the deformed space and canonical space. SMPL NN suffers from overly smooth shapes due to inaccurate correspondences when the actual shape and the skinned base mesh do not match well, \eg \ around the lapel.

\section{Conclusion}
We propose \name, a fast, robust, and universal articulation module for neural field representations. \name \ is built upon the idea of differentiable forward skinning from SNARF~\cite{Chen2021ICCV}, but is orders of magnitude faster than SNARF thanks to a series of algorithmic and implementation improvements. These include voxel-based correspondence search, LBS pre-computation, a custom CUDA kernel implementation for root finding, duplicate correspondences removal, approximated implicit gradients, and online MLP-to-voxel conversion. The resulting algorithm can find correspondences as accurately as SNARF while being $150\times$ faster. This leads to significant speed-up in various real-world applications of forward skinning algorithms. Using \name \, we are able to learn animatable human avatars from scans 15$\times$ faster than SNARF, and in contrast to SNARF, the speed bottleneck is now the canonical shape query instead of the articulation module. 
We believe \name's speed and accuracy will open new applications and accelerate research on non-rigid 3D reconstruction.

\boldparagraph{Acknowledgements} Xu Chen was supported by the Max Planck ETH Center for Learning Systems. Andreas Geiger was supported by the DFG EXC number 2064/1 - project number 390727645. We thank Yao Feng, Jinlong Yang, Yuliang Xiu and Alex Zicong Fan for their feedback.

\boldparagraph{Disclosure:} 
MJB has received research gift funds from Adobe, Intel, Nvidia, Meta/Facebook, and Amazon.  MJB has financial interests in Amazon, Datagen Technologies, and Meshcapade GmbH.  MJB's research was performed solely at, and funded solely by, the Max Planck.

\bibliographystyle{ieee_fullname}
\bibliography{bibliography_long,bibliography,bibliography_custom}

\end{document}